\pdfoutput=1
\documentclass[letterpaper]{article} 
\usepackage{aaai25}
\usepackage{times}  
\usepackage{helvet}  
\usepackage{courier}  
\usepackage[hyphens]{url}  
\usepackage{graphicx} 
\usepackage{natbib}  
\usepackage{caption} 
\usepackage{algorithm}
\usepackage{algorithmic}
\usepackage{amsmath}

\usepackage{booktabs}
\usepackage{newfloat}
\usepackage{listings}
\usepackage{xspace}
\newcommand{\firstmodule}{\emph{Dynamic Points Extraction}\xspace}
\newcommand{\secondmodule}{\emph{Moving Object Segmentation}\xspace}
\newcommand{\methodname}{\emph{MONA}}

\DeclareCaptionStyle{ruled}{labelfont=normalfont,labelsep=colon,strut=off} 
\floatstyle{ruled}
\newfloat{listing}{tb}{lst}{}
\floatname{listing}{Listing}
\def\showauthors@on{T}
\title{MONA: Moving Object Detection from Videos Shot by Dynamic Camera}
\author {
    Boxun Hu$^{*}$\textsuperscript{\rm 1},
    Mingze Xia$^{*}$\textsuperscript{\rm 1},
    Ding Zhao\textsuperscript{\rm 1},
    Guanlin Wu$^{\dag}$\textsuperscript{\rm 1}
}
\affiliations {
    \textsuperscript{\rm 1}Johns Hopkins University\\
    \{bhu29, mxia8, dzhao29, gwu32\}@jh.edu \\
    $^{*}$Equal Contribution, $^{\dag}$Corresponding Author
}

\usepackage{bibentry}
\begin{document}

\maketitle

\begin{abstract}
Dynamic urban environments, characterized by moving cameras and objects, pose significant challenges for camera trajectory estimation by complicating the distinction between camera-induced and object motion. We introduce \methodname, a novel framework designed for robust moving object detection and segmentation from videos shot by dynamic cameras. \methodname ~comprises two key modules: \firstmodule, which leverages optical flow and tracking any point to identify dynamic points, and \secondmodule, which employs adaptive bounding box filtering, and the Segment Anything for precise moving object segmentation. We validate \methodname ~by integrating with the camera trajectory estimation method LEAP-VO, and it achieves state-of-the-art results on the MPI Sintel dataset comparing to existing methods. These results demonstrate \methodname's effectiveness for moving object detection and its potential in many other applications in the urban planning field.
\end{abstract}

%

\section{Introduction}
In recent years, advancements in artificial intelligence (AI) have spurred the development of innovative systems that are transforming urban environments. Notable examples include Autonomous Driving Systems (ADS)~\cite{alex2019pointpillars, shi2019parta2, shi2019pointrcnn} and low-altitude economy system~\cite{huang2024low, li2024unauthorized}, which are reshaping urban mobility and infrastructure. Concurrently, sophisticated Human Motion Recovery (HMR) techniques~\cite{ye2023decoupling, shin2024wham, shen2024gvhmr} and human understanding methodologies~\cite{chen2024motionllm, tm2t, motiongpt} have enhanced the capabilities of human-AI collaboration in urban planning~\cite{wang2019facilitating, Nikolaos2019acyber}.


However, the complexity of urban scenarios poses significant challenges for existing methods in ADS~\cite{shi2019parta2, alex2019pointpillars}, Unmanned Aerial Vehicle (UAV) motion planning~\cite{zhou2021raptor, tordesillas2023deep, wu2024relax}, and HMR~\cite{shin2024wham, shen2024gvhmr}. These data-driven approaches often struggle to perform optimally across diverse and intricate real-world conditions. While increasing the volume of training data can enhance their performance by exposing models to a broader range of scenarios, the acquisition of marker-based datasets—datasets with precise ground truth (GT) values—is prohibitively expensive. This limitation hampers the scalability and effectiveness of such methods.

Conversely, markerless datasets, which rely on pseudo-GT annotations generated through automated pipelines applied to vast quantities of online videos, offer a cost-effective alternative for augmenting training data~\cite{lin2023motionx}. To leverage these markerless datasets effectively for applications in urban planning, it is essential to accurately estimate camera trajectories. This requirement arises because most online videos are captured using dynamic, moving cameras that do not provide GT camera trajectories.

Existing camera trajectory estimation techniques, including optical flow-based methods such as Droid-SLAM~\cite{teed2021droid} and DPVO~\cite{teed2023dpvo}, as well as Tracking Any Point (TAP) approaches like LEAP-VO~\cite{chen2024leap}, primarily rely on RGB video inputs to recover camera motion. However, their performance deteriorates in the presence of large moving objects—such as pedestrians and vehicles—that occupy significant portions of the frame. These dynamic objects can mislead the models, resulting in inaccurate trajectory estimations.

A promising strategy to mitigate this issue involves detecting and masking moving objects within the video frames during the bundle adjustment (BA) process. The primary challenge, therefore, is \textit{how to effectively detect moving objects in videos captured by dynamic cameras}. Traditional moving object detection methods, which typically depend on background subtraction and motion detection using RGB data~\cite{ellenfeld2021deep}, encounter substantial difficulties in dynamic camera scenarios. These challenges include distinguishing between camera-induced motion and object motion, managing motion blur, and handling occlusions in cluttered urban environments~\cite{yazdi2018new}.

\begin{figure*}[!t]
    \centering
    \includegraphics[width=0.8\linewidth]{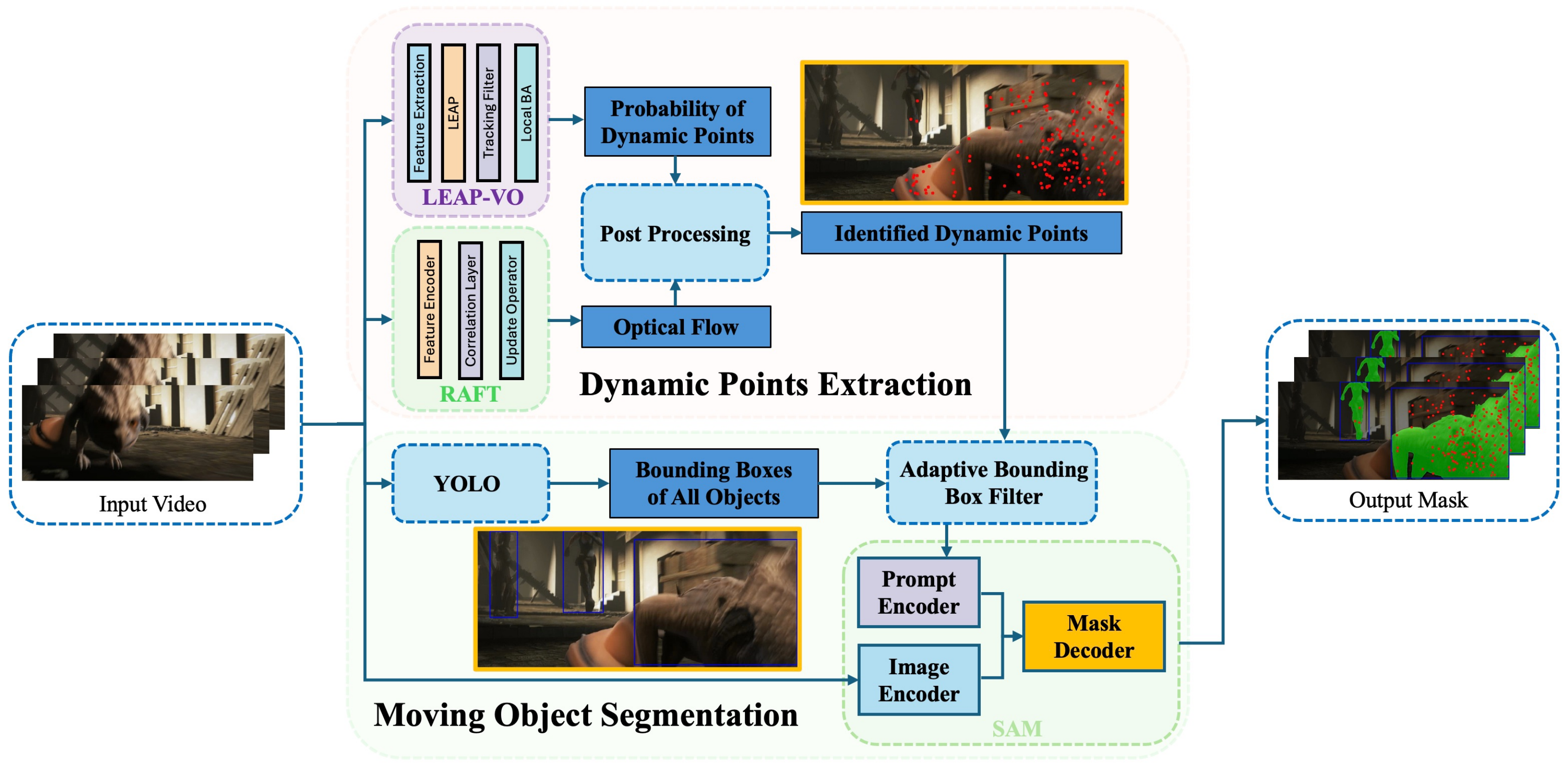}
    \caption{\textbf{The overall pipeline of \methodname.} Our method consists of two main modules: \firstmodule and \secondmodule. In \firstmodule, random points are selected and LEAP-VO estimates the probability of each point being dynamic. RAFT then computes the optical flow, and a post-processing algorithm combines these results to identify dynamic points. In \secondmodule, YOLO detects all object bounding boxes, which are filtered using the dynamic points through our designed adaptive bounding box filter to identify moving objects. Finally, the filtered bounding boxes are used as prompt input to SAM to get the segmentation of the moving objects from the input video.}
    \label{fig:Pipeline}
\end{figure*}

To address these limitations, we propose a novel framework \methodname, \underline{M}oving \underline{O}bject detection from videos shot by dy\underline{NA}mic Camera. Our approach comprises two main modules: \firstmodule and \secondmodule. The \firstmodule calculates the probability of selected points within the video are dynamic (moving) similar to LEAP-VO~\cite{chen2024leap}. Subsequently, an optical flow-based thresholding strategy is employed to classify each point as either dynamic or static. The \secondmodule leverages the identified dynamic points to segment moving objects. This is achieved by utilizing YOLO~\cite{khanam2024yolov11} to detect objects in each frame and determine the bounding boxes (bboxes) that encompass the dynamic points through our proposed adaptive bbox filtering algorithm. These bboxes serve as prompt inputs to the Segment Anything Model (SAM)~\cite{kirillov2023segment}, which generates precise masks of the moving objects. We validate our framework through downstream tasks, such as camera trajectory estimation, demonstrating state-of-the-art (SOTA) performance on the MPI Sintel Dataset~\cite{butler2012sintel}.

\section{Methods}
In this section, we introduce \methodname, a framework for detecting moving objects in videos captured by dynamic cameras. The system overview is shown in Fig.~\ref{fig:Pipeline}. Given an input video sequence $V = [F_1, \ldots, F_T]$, where $F_t$ represents the frame at time $t$, \methodname\xspace firstly uses \firstmodule to detect all dynamic points $\mathbf{X}_d$ based on LEAP-VO~\cite{chen2024leap}, RAFT~\cite{teed2020raft} and our post-processing algorithm. Subsequently, our method uses \secondmodule to detect and segment all moving objects based on the input dynamic points, YOLO~\cite{khanam2024yolov11}, SAM~\cite{kirillov2023segment} and our adaptive bounding box filter. 

\noindent\textbf{\firstmodule.} This module identifies dynamic points within the video. Following the approach of LEAP-VO~\cite{chen2024leap}, we initially select $n$ random detection points $\mathbf{X} = [\mathbf{x}_1, \ldots, \mathbf{x}_n]$ and $m$ anchor points from the first frame. Anchor points are chosen based on image gradients by dividing each frame $F_t$ into $k \times k$ grids and selecting the point with the highest gradient in each grid. We then apply the TAP method from CoTracker~\cite{karaev2024cotracker} to obtain the trajectories $a$ and visibility $v$ for both detection and anchor points, shown as
\begin{equation}
    (a, v) = \text{CoTracker}(V, \mathbf{x}_q, t_q),
    \label{eq:tap}
\end{equation}
where $\mathbf{x_q}$ is the selected point and $t_q$ is the index of the frame from which $\mathbf{x}_q$ is extracted. The image features for each frame of the video are extracted by a CNN, and we calculate point features $\mathbf{f}$ through bilinear sampling from the image feature map at query points $\mathbf{x}_q$.

\begin{table*}[!t]
    \centering
    \setlength{\tabcolsep}{20pt}
    \begin{tabular}{lccc}
    \toprule
    Method & ATE (m)$\downarrow$ & RPE trans. (m)$\downarrow$ & RPE rot. (deg)$\downarrow$ \\
    \midrule
    DROID-SLAM~\cite{teed2021droid} & 0.175 & 0.084 & 1.912 \\
    Tartan VO~\cite{tartanvo2020corl} & 0.238 & 0.084 & 1.305 \\
    Dytan VO~\cite{shen2023dytanvo} & 0.131 & 0.097 & 1.538 \\
    DPVO~\cite{teed2023dpvo} & 0.076 & 0.078 & 1.722 \\
    LEAP-VO~\cite{chen2024leap} & 0.068 & 0.035 & 0.150 \\
    \textbf{\methodname + LEAP-VO (Ours)} & \textbf{0.029} & \textbf{0.013} & \textbf{0.054} \\
    \bottomrule
    \end{tabular}%
    \caption{\textbf{Comparison of the performance between methods on MPI Sintel Dataset.} Compared to the original LEAP-VO, our proposed method (\methodname + LEAP-VO) has achieved over $60\%$ improvement on ATE, RPE trans. and RPE rot., demonstrating that \methodname's performance on moving object detection and its effectiveness of \methodname\xspace in camera trajectory estimation tasks.}
    \label{tab:mpi_sintel_comparison}
\end{table*}

%
%
Next, we compute the probability that each detection point $\mathbf{x}_i$ is dynamic by utilizing the anchor-based pixel tracking strategy from LEAP-VO. The core idea is to compare the movement patterns of detection points with those of anchor points to infer the dynamic probability of each $\mathbf{x}_i$. Let $\mathbf{x} = [x, y]$ represent the pixel coordinates, where $x$ and $y$ are assumed to be independent. The trajectory distribution of pixel $\mathbf{x}$ is modeled as the product of two univariate Cauchy distributions as
\begin{equation}
p(\mathbf{x}|V, \mathbf{x}_q) = p(x|V, \mathbf{x}_q) \cdot p(y|V, \mathbf{x}_q),     
\end{equation}
where $\mathbf{x}_q$ is the query point. The probability density function (PDF) of coordinates can be calculated as
\begin{multline}
p(x|V, \mathbf{x}_q) = \frac{\left(\frac{1}{2}\right)! \left(\frac{S + 1}{2} - 1\right)!}{\pi^{S/2} |\Sigma_x|^{1/2}} \\
\times \left[(x - \mu_x)^\top \Sigma_x^{-1} (x - \mu_x) + 1\right]^{-\frac{S + 1}{2}},
\end{multline}
where $\mu_x$ is the location matrix and $\Sigma_x$ is the scale matrix respectively, the PDF of $y$ is calculated similarly.
The scale matrices $\Sigma_x$ and $\Sigma_y$ are constructed by kernel-based estimation to ensure they are symmetric and positive definite, we apply a linear kernel to projected point features and add regularization, as shown in Eq. \ref{eq:kernel}.
\begin{equation}
    \Sigma_x = \mathbf{f}_x^\top \mathbf{f}_x + \lambda I
    \label{eq:kernel}
\end{equation}
$\lambda I$ is the regularization term. The $\Sigma_y$ can be calculated using similar procedure shown in Eq.~\ref{eq:kernel}.

After obtaining the dynamic probability $p$ for each point $\mathbf{x}$, a threshold is required to classify points as dynamic or static. Manually setting a fixed threshold is suboptimal due to varying camera movements across different frames within the same video. To address this, we propose a post-processing step that dynamically determines the threshold for each frame by integrating optical flow information. Specifically, we compute the optical flow $\mathbf{u}_t(\mathbf{x})$ for each pixel using RAFT \cite{teed2020raft} and calculate the mean magnitude $\bar{m}_t$ for frame $F_t$. This adaptive thresholding approach ensures more accurate identification of dynamic points by accounting for the specific motion characteristics of each frame. The $\bar{m}_t$ is calculated as
\begin{equation}
\bar{m}_t = \frac{1}{|\Omega|} \sum_{\mathbf{x} \in \Omega} \|\mathbf{u}_t(\mathbf{x})\|,
\end{equation}
where $\Omega$ denotes the set of all points (pixels) in the frame and $|\Omega|$ represents the total number of points. Using the mean magnitude $\bar{m}_t$, we dynamically scale the threshold $\theta_t$ for identifying dynamic points in frame $t$. Detection points $\mathbf{x}$ in frame $t$ with magnitudes exceeding $\theta_t$ are classified into the dynamic points list $\mathbf{x}_d$ and will input to \secondmodule for the segmentation of moving objects. 

\noindent\textbf{\secondmodule.} To segment moving objects from the dynamic points list $\mathbf{x}_d$, objects in each frame are first detected using YOLO, resulting in bounding boxes $\mathcal{B} = {b_1, \ldots, b_n}$. Since YOLO cannot differentiate between static and dynamic objects, an adaptive bounding box filter is applied using $\mathbf{x}_d$ and $\mathcal{B}$ to exclude bounding boxes $b_i$ corresponding to static objects. Let $D_i$ denote the number of dynamic points within bounding box $b_i$. Manually setting a threshold based on $D_i$ may inadvertently classify larger bounding boxes as moving objects. To ensure consistency across varying bounding box sizes, a threshold $\tau_0$ is established for moving objects within the video. The smallest bounding box $b_u$ in $\mathcal{B}$ where $D_u$ exceeds $\tau_0$ is identified, and its area is used as the unit area. Subsequently, the threshold $\tau^i_t$ for each bounding box $b_i$ in frame $t$ is scaled based on the ratio of the area of $b_i$ to that of $b_u$, shown as
\begin{equation}
\tau^i_t = \tau_0 \times \frac{\text{Area}(b_i)}{\text{Area}(b_u)}, u = \arg \min_{b \in \mathcal{B}} \left\{ \text{Area}(b) \mid D_u \geq \tau_0 \right\}
\end{equation}
The filtered bounding boxes are defined as $\mathcal{B}_{\text{filtered}} = \left\{ b_i \mid D_i \geq \tau^i_t, \; b_i \in \mathcal{B} \right\}$,
which determine whether the bounding boxes belong to truly moving objects. $\mathcal{B}_\text{filtered}$ denotes high-quality prompts to SAM for segmentation.

\section{Experiments and Results}
Currently, no public dataset is available to directly evaluate detection accuracy and mask quality for moving objects in videos captured by moving cameras. Therefore, we selected camera trajectory estimation as our downstream testing task, based on the hypothesis that accurately detecting and masking moving objects enhances camera trajectory estimation. By masking dynamic objects, the model can focus on static regions, improving the selection of tracked points and reducing noise from moving entities. For this purpose, we integrate our method with LEAP-VO \cite{chen2024leap}, a camera trajectory estimation method, to evaluate \methodname's performance for moving object detection by analyzing the output estimated camera trajectory of \methodname+LEAP-VO.

\begin{figure}
    \centering
    \includegraphics[width=\linewidth]{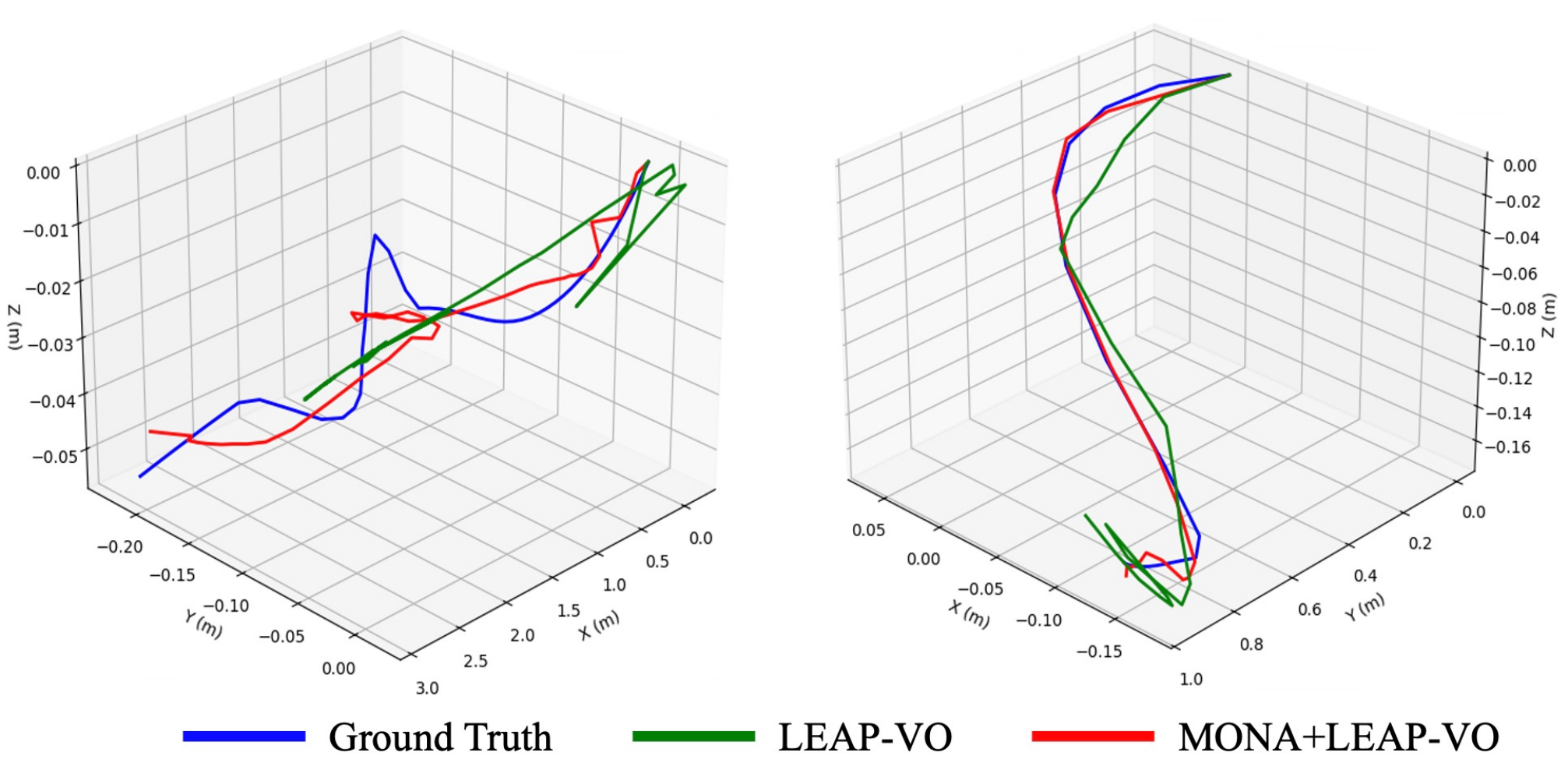}
    \caption{\textbf{Qualitative Comparison on Estimated Camera Trajectory between different methods.} We run original LEAP-VO and LEAP-VO with \methodname\xspace on MPI Sintel Dataset~\cite{butler2012sintel}. Two estimated trajectories are selected as the comparison. The \methodname+LEAP-VO provides a more accurate estimated trajectory as it is more similar to the GT, which demonstrates the effectiveness of our method in camera trajectory estimation tasks.}
    \label{fig:Traj_compare}
\end{figure}

For an initial qualitative evaluation, we compared the original LEAP-VO, \methodname+LEAP-VO, and the ground truth trajectories using raw videos from the MPI Sintel dataset~\cite{butler2012sintel}. Figure~\ref{fig:Traj_compare} shows two selected examples, demonstrating that our method significantly enhances trajectory estimation. By accurately detecting and masking moving objects during bundle adjustment in LEAP-VO, \methodname+LEAP-VO produces camera trajectories that closely match the ground truth, regardless of the complexity of the camera motion. These results highlight the robustness and adaptability of our approach across various scenarios.

To quantitatively evaluate our proposed method, we compared the original LEAP-VO with \methodname+LEAP-VO on the MPI Sintel dataset \cite{butler2012sintel} for camera trajectory estimation. As shown in Table \ref{tab:mpi_sintel_comparison}, we assessed performance using three metrics: Absolute Trajectory Error (ATE), Relative Translation Error (RPE trans), and Relative Rotation Error (RPE rot). \methodname+LEAP-VO achieved over a 60\% improvement across all metrics, significantly outperforming existing SOTA methods. These results validate the effectiveness of our approach in enhancing trajectory estimation accuracy, particularly in scenarios with moving objects.

The substantial improvements of \methodname+LEAP-VO in camera trajectory estimation are due to its enhanced selection of tracked points in the TAP pipeline through moving object detection and segmentation. While LEAP-VO initially selects random points and filters dynamic ones by comparing them to anchor points, this approach does not consistently avoid dynamic object regions due to the difficulty in determining optimal thresholds. By incorporating \methodname, it can address this limitation by first detecting all moving objects and getting their masks. Then, the randomly selected points inside the moving objects mask can be filtered in the bundle adjustment process in LEAP-VO, thereby improving the quality of output estimated camera trajectory.

\begin{figure}[ht]
    \centering
    \includegraphics[width=1\linewidth]{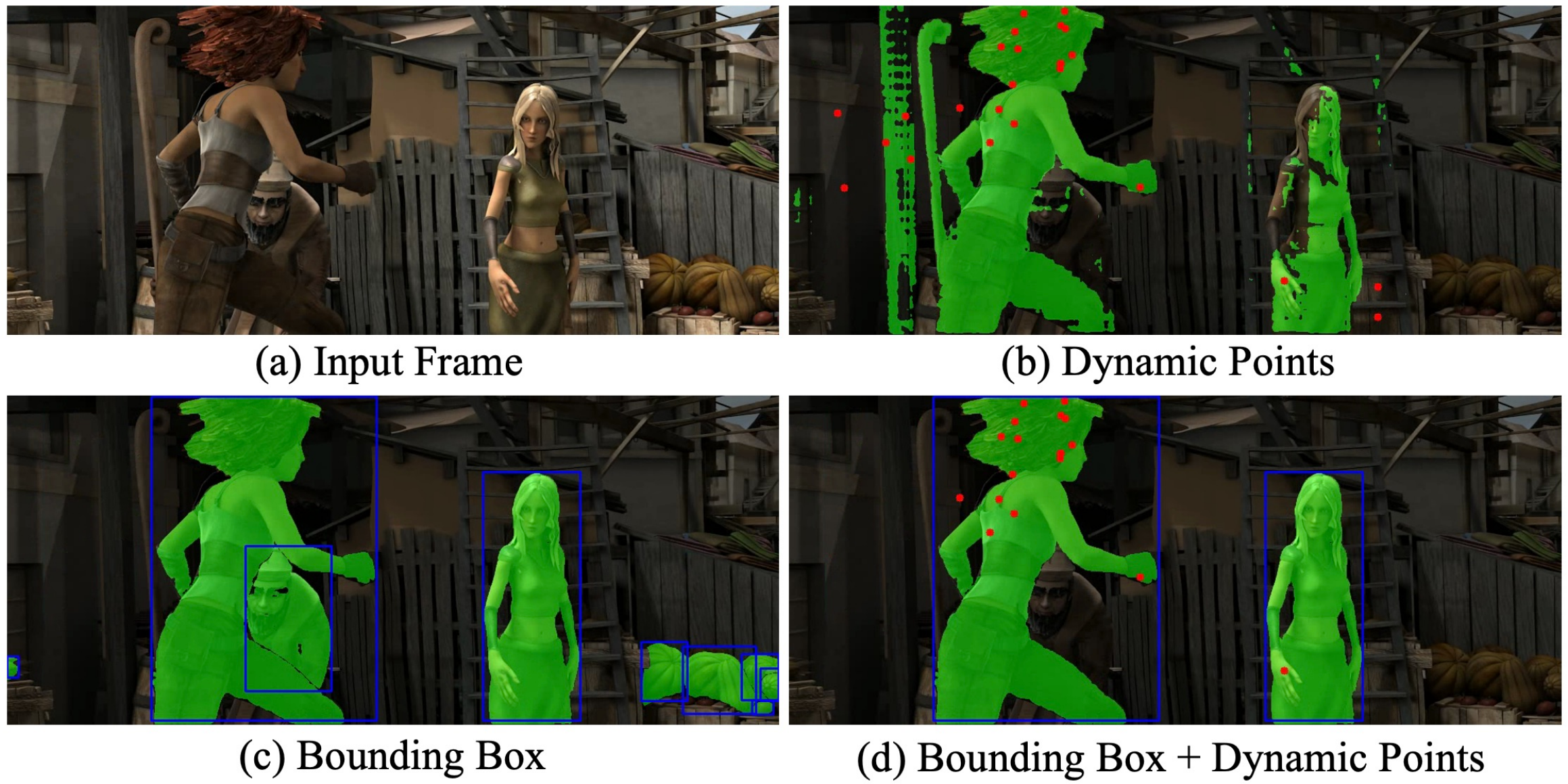}
    \caption{\textbf{Qualitative Ablation Study of \methodname.} We visualize and compare the quality of masks produced by different prompts input to SAM: (b) pure dynamic points (c) pure bounding boxes (without filtering), and (d) bounding boxes with dynamic points filtering strategy.}
    \label{fig:Mask_comparison}
\end{figure}

\noindent\textbf{Ablation Study.} We compared the segmentation results produced by three approaches: using only dynamic points as prompts, using YOLO’s raw bounding boxes, and using our filtered bounding boxes. The visualization of Fig.~\ref{fig:Mask_comparison} visualizes these results. Using only dynamic points as prompts resulted in incomplete masks and incorrect segmentations (Fig.~\ref{fig:Mask_comparison} (b)), while YOLO's raw bounding boxes failed to distinguish between static and dynamic objects ((Fig.~\ref{fig:Mask_comparison} (c))). In contrast, our pipeline, which incorporates filtered bounding boxes, consistently produced accurate and effective segmentation masks for moving objects (Fig.~\ref{fig:Mask_comparison} (d)). These findings demonstrate the superiority of our approach in accurately detecting moving objects and generating high-quality masks, which are essential for improving the performance of downstream tasks like camera trajectory estimation.

\section{Conclusion}
In this paper, we introduce \methodname, a robust framework for detecting moving objects in RGB videos captured by dynamic cameras. Our method effectively separates camera-induced motion from object motion in in-the-wild footage. \methodname\xspace enhances various tasks, particularly camera trajectory estimation, by complementing existing approaches. When integrated with LEAP-VO, \methodname\xspace achieves state-of-the-art performance on the MPI Sintel Dataset. Given that an accurate camera trajectory estimation is essential for creating markerless datasets used in ADS, UAV motion planning, and HMR, \methodname\xspace offers new opportunities in both datasets and other applications in the urban planning domain.

\bibliography{aaai25}

\begin{thebibliography}{46}
\providecommand{\natexlab}[1]{#1}

\bibitem[{Bescos et~al.(2018)Bescos, Fácil, Civera, and Neira}]{bescos2018dynaslam}
Bescos, B.; Fácil, J.~M.; Civera, J.; and Neira, J. 2018.
\newblock DynaSLAM: Tracking, Mapping, and Inpainting in Dynamic Scenes.
\newblock \emph{IEEE Robotics and Automation Letters}, 3(4): 4076--4083.

\bibitem[{Butler et~al.(2012)Butler, Wulff, Stanley, and Black}]{butler2012sintel}
Butler, D.~J.; Wulff, J.; Stanley, G.~B.; and Black, M.~J. 2012.
\newblock A naturalistic open source movie for optical flow evaluation.
\newblock In {A. Fitzgibbon et al. (Eds.)}, ed., \emph{European Conf. on Computer Vision (ECCV)}, Part IV, LNCS 7577, 611--625. Springer-Verlag.

\bibitem[{Caldeira et~al.(2020)Caldeira, Fout, Kesari, Sefala, Walsh, Dupre, Khaefi, Setiaji, Hodge, Pramestri et~al.}]{caldeira2020improving}
Caldeira, J.; Fout, A.; Kesari, A.; Sefala, R.; Walsh, J.; Dupre, K.; Khaefi, M.~R.; Setiaji; Hodge, G.; Pramestri, Z.~A.; et~al. 2020.
\newblock Improving traffic safety through video analysis in Jakarta, Indonesia.
\newblock In \emph{Intelligent Systems and Applications: Proceedings of the 2019 Intelligent Systems Conference (IntelliSys) Volume 2}, 642--649. Springer.

\bibitem[{Chapel and Bouwmans(2020)}]{chapel2020moving}
Chapel, M.-N.; and Bouwmans, T. 2020.
\newblock Moving objects detection with a moving camera: A comprehensive review.
\newblock \emph{Computer science review}, 38: 100310.

\bibitem[{Chen et~al.(2024{\natexlab{a}})Chen, Lu, Zeng, Zhang, Wang, Zhang, and Zhang}]{chen2024motionllm}
Chen, L.-H.; Lu, S.; Zeng, A.; Zhang, H.; Wang, B.; Zhang, R.; and Zhang, L. 2024{\natexlab{a}}.
\newblock MotionLLM: Understanding Human Behaviors from Human Motions and Videos.
\newblock \emph{arXiv preprint arXiv:2405.20340}.

\bibitem[{Chen et~al.(2024{\natexlab{b}})Chen, Chen, Wang, and Pollefeys}]{chen2024leap}
Chen, W.; Chen, L.; Wang, R.; and Pollefeys, M. 2024{\natexlab{b}}.
\newblock LEAP-VO: Long-term Effective Any Point Tracking for Visual Odometry.
\newblock In \emph{CVPR}.

\bibitem[{Chen et~al.(2017)Chen, Ma, Wan, Li, and Xia}]{chen2017multiview}
Chen, X.; Ma, H.; Wan, J.; Li, B.; and Xia, T. 2017.
\newblock Multi-view 3D Object Detection Network for Autonomous Driving.
\newblock In \emph{2017 IEEE Conference on Computer Vision and Pattern Recognition (CVPR)}, 6526--6534.

\bibitem[{Chen et~al.(2019)Chen, Liu, Shen, and Jia}]{chen2019fastprcnn}
Chen, Y.; Liu, S.; Shen, X.; and Jia, J. 2019.
\newblock Fast Point R-CNN.
\newblock In \emph{2019 IEEE/CVF International Conference on Computer Vision (ICCV)}, 9774--9783.

\bibitem[{Ellenfeld et~al.(2021)Ellenfeld, Moosbauer, Cardenes, Klauck, and Teutsch}]{ellenfeld2021deep}
Ellenfeld, M.; Moosbauer, S.; Cardenes, R.; Klauck, U.; and Teutsch, M. 2021.
\newblock Deep fusion of appearance and frame differencing for motion segmentation.
\newblock In \emph{proceedings of the IEEE/CVF Conference on Computer Vision and Pattern Recognition}, 4339--4349.

\bibitem[{Foth et~al.(2014)Foth, Heikkinen, Ylipulli, Luusua, Satchell, and Ojala}]{foth2014ubiopticon}
Foth, M.; Heikkinen, T.; Ylipulli, J.; Luusua, A.; Satchell, C.; and Ojala, T. 2014.
\newblock UbiOpticon: participatory sousveillance with urban screens and mobile phone cameras.
\newblock In \emph{Proceedings of The International Symposium on Pervasive Displays}, 56--61.

\bibitem[{Guo et~al.(2022)Guo, Zuo, Wang, and Cheng}]{tm2t}
Guo, C.; Zuo, X.; Wang, S.; and Cheng, L. 2022.
\newblock Tm2t: Stochastic and tokenized modeling for the reciprocal generation of 3d human motions and texts.
\newblock In \emph{ECCV}, 580--597.

\bibitem[{Huang et~al.(2024)Huang, Fang, Wu, Wang, and Yang}]{huang2024low}
Huang, C.; Fang, S.; Wu, H.; Wang, Y.; and Yang, Y. 2024.
\newblock Low-altitude intelligent transportation: System architecture, infrastructure, and key technologies.
\newblock \emph{Journal of Industrial Information Integration}, 42: 100694.

\bibitem[{Jiang et~al.(2024)Jiang, Chen, Liu, Yu, Yu, and Chen}]{motiongpt}
Jiang, B.; Chen, X.; Liu, W.; Yu, J.; Yu, G.; and Chen, T. 2024.
\newblock Motiongpt: Human motion as a foreign language.
\newblock \emph{NeurIPS}.

\bibitem[{Karaev et~al.(2024)Karaev, Rocco, Graham, Neverova, Vedaldi, and Rupprecht}]{karaev2024cotracker}
Karaev, N.; Rocco, I.; Graham, B.; Neverova, N.; Vedaldi, A.; and Rupprecht, C. 2024.
\newblock CoTracker: It is Better to Track Together.
\newblock In \emph{Proc. {ECCV}}.

\bibitem[{Karagulian et~al.(2023)Karagulian, Liberto, Corazza, Valenti, Dumitru, and Nigro}]{karagulian2023pedestrian}
Karagulian, F.; Liberto, C.; Corazza, M.; Valenti, G.; Dumitru, A.; and Nigro, M. 2023.
\newblock Pedestrian flows characterization and estimation with computer vision techniques.
\newblock \emph{Urban Science}, 7(2): 65.

\bibitem[{Khanam and Hussain(2024)}]{khanam2024yolov11}
Khanam, R.; and Hussain, M. 2024.
\newblock Yolov11: An overview of the key architectural enhancements.
\newblock \emph{arXiv preprint arXiv:2410.17725}.

\bibitem[{Kirillov et~al.(2023)Kirillov, Mintun, Ravi, Mao, Rolland, Gustafson, Xiao, Whitehead, Berg, Lo et~al.}]{kirillov2023segment}
Kirillov, A.; Mintun, E.; Ravi, N.; Mao, H.; Rolland, C.; Gustafson, L.; Xiao, T.; Whitehead, S.; Berg, A.~C.; Lo, W.-Y.; et~al. 2023.
\newblock Segment anything.
\newblock In \emph{Proceedings of the IEEE/CVF International Conference on Computer Vision}, 4015--4026.

\bibitem[{Ku et~al.(2018)Ku, Mozifian, Lee, Harakeh, and Waslander}]{jason2018joint}
Ku, J.; Mozifian, M.; Lee, J.; Harakeh, A.; and Waslander, S.~L. 2018.
\newblock Joint 3D Proposal Generation and Object Detection from View Aggregation.
\newblock In \emph{2018 IEEE/RSJ International Conference on Intelligent Robots and Systems (IROS)}, 1--8.

\bibitem[{Lang et~al.(2019)Lang, Vora, Caesar, Zhou, Yang, and Beijbom}]{alex2019pointpillars}
Lang, A.~H.; Vora, S.; Caesar, H.; Zhou, L.; Yang, J.; and Beijbom, O. 2019.
\newblock PointPillars: Fast Encoders for Object Detection From Point Clouds.
\newblock In \emph{2019 IEEE/CVF Conference on Computer Vision and Pattern Recognition (CVPR)}, 12689--12697.

\bibitem[{Li(2017)}]{li20173d}
Li, B. 2017.
\newblock 3D fully convolutional network for vehicle detection in point cloud.
\newblock In \emph{2017 IEEE/RSJ International Conference on Intelligent Robots and Systems (IROS)}, 1513--1518.

\bibitem[{Li et~al.(2019)Li, Ouyang, Sheng, Zeng, and Wang}]{li2019gs3d}
Li, B.; Ouyang, W.; Sheng, L.; Zeng, X.; and Wang, X. 2019.
\newblock GS3D: An Efficient 3D Object Detection Framework for Autonomous Driving.
\newblock 1019--1028.

\bibitem[{Li et~al.(2024)Li, Gao, Wang, Mei, Zhu, Chen, Wu, and Niyato}]{li2024unauthorized}
Li, Z.; Gao, Z.; Wang, K.; Mei, Y.; Zhu, C.; Chen, L.; Wu, X.; and Niyato, D. 2024.
\newblock Unauthorized UAV Countermeasure for Low-Altitude Economy: Joint Communications and Jamming Based on MIMO Cellular Systems.
\newblock \emph{IEEE Internet of Things Journal}, 1--1.

\bibitem[{Lin et~al.(2023)Lin, Zeng, Lu, Cai, Zhang, Wang, and Zhang}]{lin2023motionx}
Lin, J.; Zeng, A.; Lu, S.; Cai, Y.; Zhang, R.; Wang, H.; and Zhang, L. 2023.
\newblock Motion-X: A Large-scale 3D Expressive Whole-body Human Motion Dataset.
\newblock \emph{Advances in Neural Information Processing Systems}.

\bibitem[{Moon, Choi, and Lee(2022)}]{moon2022neuralannot}
Moon, G.; Choi, H.; and Lee, K.~M. 2022.
\newblock NeuralAnnot: Neural Annotator for 3D Human Mesh Training Sets.
\newblock In \emph{Computer Vision and Pattern Recognition Workshop (CVPRW)}.

\bibitem[{Mur-Artal and Tard\'os(2015)}]{mur2015orbslam}
Mur-Artal, M. J. M.~M., Ra\'ul; and Tard\'os, J.~D. 2015.
\newblock {ORB-SLAM}: a Versatile and Accurate Monocular {SLAM} System.
\newblock \emph{IEEE Transactions on Robotics}, 31(5): 1147--1163.

\bibitem[{Myagmar-Ochir and Kim(2023)}]{myagmar2023survey}
Myagmar-Ochir, Y.; and Kim, W. 2023.
\newblock A survey of video surveillance systems in smart city.
\newblock \emph{Electronics}, 12(17): 3567.

\bibitem[{Nikolakis, Maratos, and Makris(2019)}]{Nikolaos2019acyber}
Nikolakis, N.; Maratos, V.; and Makris, S. 2019.
\newblock A cyber physical system (CPS) approach for safe human-robot collaboration in a shared workplace.
\newblock \emph{Robotics and Computer-Integrated Manufacturing}, 56: 233--243.

\bibitem[{Pang et~al.(2024)Pang, Cai, Yang, Zhang, and Liu}]{peng2024benchmarking}
Pang, H.~E.; Cai, Z.; Yang, L.; Zhang, T.; and Liu, Z. 2024.
\newblock Benchmarking and analyzing 3D human pose and shape estimation beyond algorithms.
\newblock In \emph{Proceedings of the 36th International Conference on Neural Information Processing Systems}, NIPS '22. Red Hook, NY, USA: Curran Associates Inc.
\newblock ISBN 9781713871088.

\bibitem[{Shen et~al.(2023)Shen, Cai, Wang, and Scherer}]{shen2023dytanvo}
Shen, S.; Cai, Y.; Wang, W.; and Scherer, S. 2023.
\newblock DytanVO: Joint Refinement of Visual Odometry and Motion Segmentation in Dynamic Environments.
\newblock In \emph{2023 IEEE International Conference on Robotics and Automation (ICRA)}, 4048--4055.

\bibitem[{Shen et~al.(2024)Shen, Pi, Xia, Cen, Peng, Hu, Bao, Hu, and Zhou}]{shen2024gvhmr}
Shen, Z.; Pi, H.; Xia, Y.; Cen, Z.; Peng, S.; Hu, Z.; Bao, H.; Hu, R.; and Zhou, X. 2024.
\newblock World-Grounded Human Motion Recovery via Gravity-View Coordinates.
\newblock In \emph{ACM SIGGRAPH Asia}.

\bibitem[{Shi, Wang, and Li(2019)}]{shi2019pointrcnn}
Shi, S.; Wang, X.; and Li, H. 2019.
\newblock PointRCNN: 3D Object Proposal Generation and Detection From Point Cloud.
\newblock In \emph{The IEEE Conference on Computer Vision and Pattern Recognition (CVPR)}.

\bibitem[{Shi et~al.(2019)Shi, Wang, Shi, Wang, and Li}]{shi2019parta2}
Shi, S.; Wang, Z.; Shi, J.; Wang, X.; and Li, H. 2019.
\newblock From Points to Parts: 3D Object Detection from Point Cloud with Part-aware and Part-aggregation Network.
\newblock \emph{arXiv preprint arXiv:1907.03670}.

\bibitem[{Shin et~al.(2024)Shin, Kim, Halilaj, and Black}]{shin2024wham}
Shin, S.; Kim, J.; Halilaj, E.; and Black, M.~J. 2024.
\newblock {WHAM}: Reconstructing World-grounded Humans with Accurate {3D} Motion.
\newblock In \emph{CVPR}.

\bibitem[{Teed and Deng(2020)}]{teed2020raft}
Teed, Z.; and Deng, J. 2020.
\newblock Raft: Recurrent all-pairs field transforms for optical flow.
\newblock In \emph{Computer Vision--ECCV 2020: 16th European Conference, Glasgow, UK, August 23--28, 2020, Proceedings, Part II 16}, 402--419. Springer.

\bibitem[{Teed and Deng(2021)}]{teed2021droid}
Teed, Z.; and Deng, J. 2021.
\newblock DROID-SLAM: Deep Visual SLAM for Monocular, Stereo, and RGB-D Cameras.
\newblock In Ranzato, M.; Beygelzimer, A.; Dauphin, Y.; Liang, P.; and Vaughan, J.~W., eds., \emph{NeurIPS}, volume~34, 16558--16569. Curran Associates, Inc.

\bibitem[{Teed, Lipson, and Deng(2023)}]{teed2023dpvo}
Teed, Z.; Lipson, L.; and Deng, J. 2023.
\newblock Deep Patch Visual Odometry.
\newblock \emph{NeurIPS}.

\bibitem[{Tordesillas and How(2023)}]{tordesillas2023deep}
Tordesillas, J.; and How, J.~P. 2023.
\newblock Deep-PANTHER: Learning-Based Perception-Aware Trajectory Planner in Dynamic Environments.
\newblock \emph{IEEE Robotics and Automation Letters}, 8(3): 1399--1406.

\bibitem[{Wang et~al.(2023)Wang, Yuan, Luo, Xie, Lin, Iqbal, Fidler, and Khamis}]{wang2023learning}
Wang, J.; Yuan, Y.; Luo, Z.; Xie, K.; Lin, D.; Iqbal, U.; Fidler, S.; and Khamis, S. 2023.
\newblock Learning Human Dynamics in Autonomous Driving Scenarios.
\newblock In \emph{ICCV}, 20739--20749.

\bibitem[{Wang, Hu, and Scherer(2020)}]{tartanvo2020corl}
Wang, W.; Hu, Y.; and Scherer, S. 2020.
\newblock TartanVO: A Generalizable Learning-based VO.

\bibitem[{Wang et~al.(2019)Wang, Li, Chen, Diekel, and Jia}]{wang2019facilitating}
Wang, W.; Li, R.; Chen, Y.; Diekel, Z.~M.; and Jia, Y. 2019.
\newblock Facilitating Human–Robot Collaborative Tasks by Teaching-Learning-Collaboration From Human Demonstrations.
\newblock \emph{IEEE Transactions on Automation Science and Engineering}, 16(2): 640--653.

\bibitem[{Wu, Zhao, and He(2024)}]{wu2024relax}
Wu, G.; Zhao, Z.; and He, Y. 2024.
\newblock RELAX: Reinforcement Learning Enabled 2D-LiDAR Autonomous System for Parsimonious UAVs.
\newblock arXiv:2309.08095.

\bibitem[{Xiong, Xie, and Leng(2024)}]{xiong2024evtol}
Xiong, K.; Xie, J.; and Leng, S. 2024.
\newblock eVTOL Communication and Trajectory Optimization in Low-Altitude Economy.
\newblock In \emph{2024 IEEE/CIC International Conference on Communications in China (ICCC Workshops)}, 845--850.

\bibitem[{Yazdi and Bouwmans(2018)}]{yazdi2018new}
Yazdi, M.; and Bouwmans, T. 2018.
\newblock New trends on moving object detection in video images captured by a moving camera: A survey.
\newblock \emph{Computer science review}, 28: 157--177.

\bibitem[{Ye et~al.(2023)Ye, Pavlakos, Malik, and Kanazawa}]{ye2023decoupling}
Ye, V.; Pavlakos, G.; Malik, J.; and Kanazawa, A. 2023.
\newblock Decoupling Human and Camera Motion from Videos in the Wild.
\newblock In \emph{CVPR}.

\bibitem[{Yuan et~al.(2022)Yuan, Iqbal, Molchanov, Kitani, and Kautz}]{yuan2022glamr}
Yuan, Y.; Iqbal, U.; Molchanov, P.; Kitani, K.; and Kautz, J. 2022.
\newblock Glamr: Global occlusion-aware human mesh recovery with dynamic cameras.
\newblock In \emph{Proceedings of the IEEE/CVF conference on computer vision and pattern recognition}, 11038--11049.

\bibitem[{Zhou et~al.(2021)Zhou, Pan, Gao, and Shen}]{zhou2021raptor}
Zhou, B.; Pan, J.; Gao, F.; and Shen, S. 2021.
\newblock RAPTOR: Robust and Perception-Aware Trajectory Replanning for Quadrotor Fast Flight.
\newblock \emph{IEEE Transactions on Robotics}, 37(6): 1992--2009.

\end{thebibliography}

\end{document}